\documentclass{IEEEtran}
\usepackage{cite}

\usepackage{graphicx}
\usepackage{amsmath,amssymb} 
\usepackage{color}
 \usepackage{bbm, dsfont}
\usepackage{multirow}
 
\usepackage{longtable}
\usepackage[T1]{fontenc}
\usepackage{epsfig}
\usepackage{authblk}
 
\usepackage{floatrow}
\usepackage{caption}
\usepackage{hyperref}

\usepackage{graphicx}
\usepackage{comment}
\usepackage{amsmath,amssymb} 
\usepackage{color}
\usepackage{subfigure}
\usepackage[]{animate}
\usepackage{animate}
\usepackage{float}
\usepackage{bm}
\usepackage{multirow}
\usepackage{rotating}
\usepackage{wrapfig}
\usepackage{lineno}
\usepackage{epsfig}
\usepackage{nicefrac}       
\usepackage{microtype} 
\usepackage{float}
\usepackage{amsmath}
\usepackage{enumerate}
\usepackage{enumitem} 
\usepackage{soul}
\usepackage{graphicx}
\usepackage{amsmath,amssymb} 
 
\usepackage{color}

\usepackage[noend]{algpseudocode}
\usepackage{multirow}
 
\usepackage{longtable}
\usepackage[T1]{fontenc}
\usepackage{epsfig}

\usepackage{floatrow}
\usepackage{caption}

\usepackage{url}

\usepackage{hyperref}

\usepackage{mathtools}
\usepackage{resizegather}
\usepackage{dsfont}
\usepackage{caption}
\usepackage{multirow}

\usepackage{amsthm}
\theoremstyle{plain}

\usepackage[section]{placeins}

\usepackage{amsmath,amssymb,amsfonts}
\usepackage{graphicx}
\usepackage{textcomp,nicefrac}
\def\BibTeX{{\rm B\kern-.05em{\sc i\kern-.025em b}\kern-.08em
T\kern-.1667em\lower.7ex\hbox{E}\kern-.125emX}}
\begin{document}
\title{Point-supervised Brain Tumor Segmentation with Box-prompted MedSAM}
\author{Xiaofeng Liu, Jonghye Woo, Chao Ma, Jinsong Ouyang, and Georges El Fakhri
\thanks{X. Liu, C. Ma, J. Ouyang and G. El Fakhri are with the Department of Radiology and Biomedical Imaging, Yale University, New Haven, CT 06519.}
\thanks{J. Woo is with the Massachusetts General Hospital and Harvard Medical School, Boston, MA 02114.}
}

\maketitle

\begin{abstract}
Delineating lesions and anatomical structure is important for image-guided interventions. Point-supervised medical image segmentation (PSS) has great potential to alleviate costly expert delineation labeling. However, due to the lack of precise size and boundary guidance, the effectiveness of PSS often falls short of expectations. Although recent vision foundational models, such as the medical segment anything model (MedSAM), have made significant advancements in bounding-box-prompted segmentation, it is not straightforward to utilize point annotation, and is prone to semantic ambiguity. In this preliminary study, we introduce an iterative framework to facilitate semantic-aware point-supervised MedSAM. Specifically, the semantic box-prompt generator (SBPG) module has the capacity to convert the point input into potential pseudo bounding box suggestions, which are explicitly refined by the prototype-based semantic similarity. This is then succeeded by a prompt-guided spatial refinement (PGSR) module that harnesses the exceptional generalizability of MedSAM to infer the segmentation mask, which also updates the box proposal seed in SBPG. Performance can be progressively improved with adequate iterations. We conducted an evaluation on BraTS2018 for the segmentation of whole brain tumors and demonstrated its superior performance compared to traditional PSS methods and on par with box-supervised methods.
\end{abstract}

\vspace{-4pt}
\section{Introduction} 
\label{sec:introduction}

\IEEEPARstart{T}{he} costly labeling effort in medical image delineation significantly hinder the development of data-driven AI models. There are increasing interests on weakly supervised segmentation to utilize bounding box or even a single point as label for training supervision~\cite{shen2023survey}. However, due to the absence of accurate size and boundary guidance, there is a large performance gap between point-supervised medical image segmentation (PSS) and mask/box-supervised counterparts~\cite{guo2023p2p}. 

The recent progress of vision foundational models has achieved breakthroughs in several weakly supervised tasks, benefited by their strong zero-shot generalizability. For instance, the point-prompt natural image segment anything model (SAM) has been applied to enhance pseudo labels for point-supervised segmentation~\cite{chen2023segment,guo2023p2p}. Recently, the medical image version of SAM (MedSAM)~\cite{ma2024segment} has been trained with 1.5 million segmented medical images to achieve generalizable \textit{box-prompt} segmentation. MedSAM takes both an image slice and a bounding box, i.e., prompt, to predict the possible segmentation mask within the box. Notably, the natural image SAM itself follows \textit{point-prompt} design. Therefore, it is not straight forward to integrate box-prompt MedSAM to PSS task as~\cite{chen2023segment}. In addition, a significant limitation of SAM/MedSAM is the lack of classification ability, resulting in class / structure-agnostic segmentation results~\cite{chen2023segment,guo2023p2p}. They are designed for general use, while fails to accurately delineate the specific lesion or structure as desired~\cite{chen2023segment,guo2023p2p}.

\begin{figure}[!t]
\begin{center}
\includegraphics[width=1\linewidth]{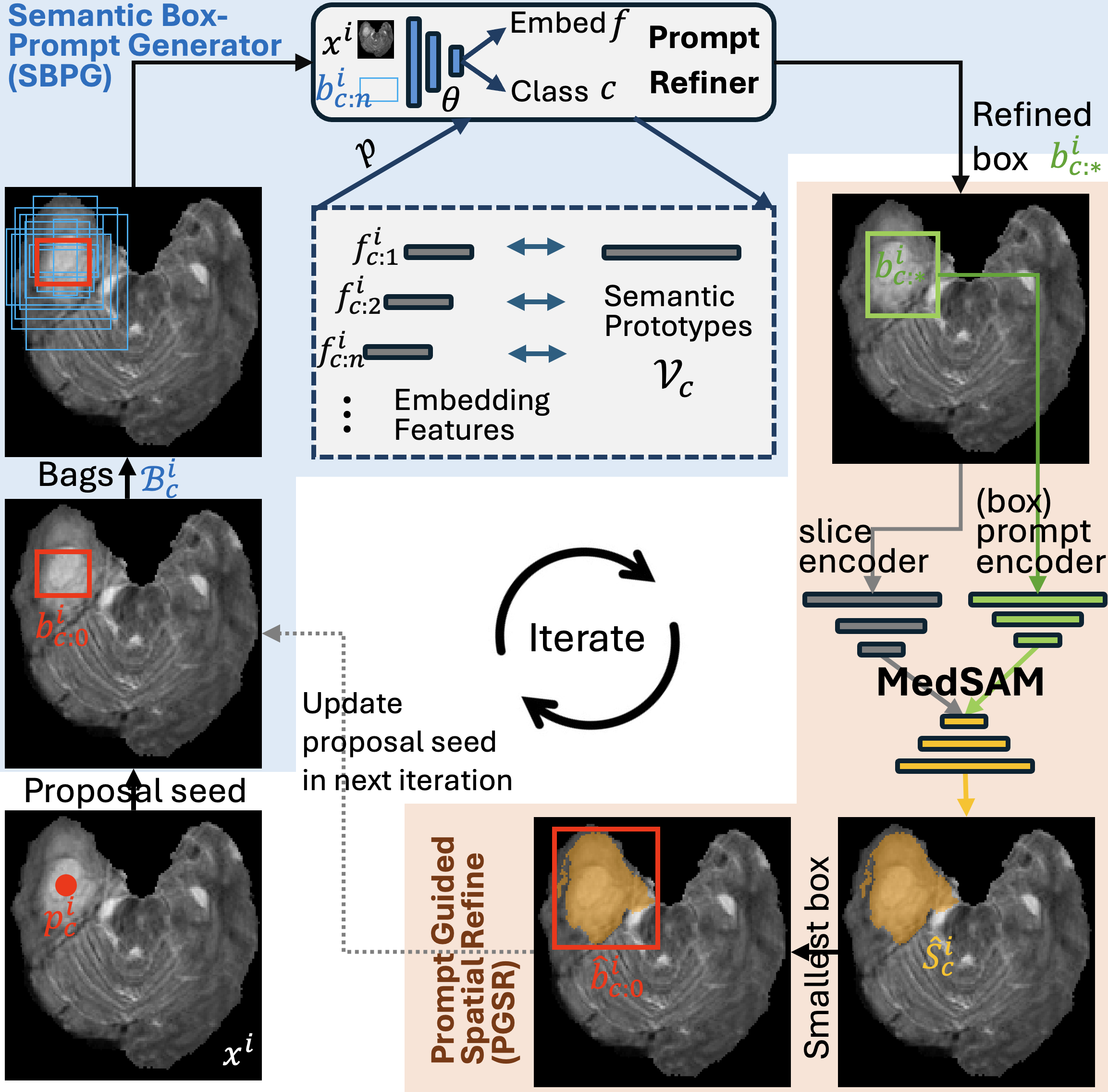}    
\end{center} 
\caption{Proposed iterative refinement framework with SBPG and PGSR modules for semantic-aware PSS utilizing the off-the-shelf box-prompt MedSAM. Only prompt refiner $\theta$ is to be trained with point-supervision.}
\label{fig:illus} \vspace{-15pt}
\end{figure} 

To our knowledge, this is the first attempt to integrate MedSAM to facilitate semantic-aware PSS. We adopt an iterative framework to achieve coarse-to-fine progression of the bounding box. The point prompt is first converted to a box-proposal seed. Then, we configure a semantic box-prompt generator (SBPG) to propose and pick the reasonable box according to semantic similarity as~\cite{guo2023p2p}. It is followed by a box prompt guided spatial refinement (PGSR) to utilize the generalizable MedSAM to predict the segmentation mask. In addition, the smallest box that covers the mask is further used as the box seed in the next round of SBPG. We do not need to fine-tune the MedSAM to be semantic aware, which is prone to catastrophic forgetting. Notably, only the prompt refiner $\theta$ with the encoder part of the segmentor model will be trained, which involves about half the parameters of mask-supervised UNet training. 

We demonstrated its effectiveness in BraTS2018 for PSS of the whole brain tumor segmentation from T2-weighted MRI slices. We show that in testing, with 3 to 5 rounds of iteration, the point-prompt can achieve superior performance to approximate the box-supervised MedSAM~\cite{ma2024segment}.

\vspace{-4pt}
\section{Methodology}
In PSS, we are given tuplet $\{x^i,\rho_c^i,s^i\}$, in which $x^i$ can be an MR slice, while the point prompt $\rho^i_c=\{\rho^i_x,\rho^i_y,c\}$ indicates the 2D spatial coordinates and the interested class $c$, e.g., tumor or normal tissue. We would expect the segmentation mask prediction $\hat{s}^i_c$ to approximate the expert annotation $s^i_c$. 

To catering the box-prompt MedSAM~\cite{ma2024segment}, an initial box proposal seed of class $c$, i.e., $b_{c:0}^i$, is generated from $\rho^i_c$, which is centered on point location $(\rho^i_x,\rho^i_y)$ with the size of $X\times Y$. We do not expect $b_{c:0}^i$ fit the region of interests very well as there is no prior information about the size in the initial step. Then, the proposal bag $\mathcal{B}^i_c=\{b_{c:n}^i\}_{n=1}^N$ is created by scaling $b_{c:0}^i$ with $N$ different scales. Therefore, the boxes in group $\mathcal{B}^i_c$ has strong spatial correlation with $\rho_c^i$, which avoid the inefficient random proposal in whole image~\cite{guo2023p2p}. 

To enable the model be aware of the specific semantic class, for example tumor in our task, a parameterized prompt refiner $\theta$ with a segmentor encoder and two fully connected layers is applied. Specifically, for the training sample with segmentation mask label $s^i_c$, we convert $s^i_c$ to the corresponding smallest rectangle bounding box label $b^i_c$. Then, we store a set of feature $f^i_{c:n}=\theta(x^i,b_{c:n}^i)$ in the most recent $M$ batches to a memory buffer, and calculate the mean of $\{f^i_{c:nm}\}_{n,m=1}^{N,M}$ as the prototype $\mathcal{V}_c$. As conventional multiple instance learning (MIL) based methods~\cite{shen2023survey},
the instance level probability indicating the likelihood of $b^i_{c:n}$ matches $\mathcal{V}_c$ can be approximated with {\begin{equation}\vspace{-12pt}
\begin{aligned}
p(b^i_{c:n},\mathcal{V}_c)=\frac{e^{cos(f^i_{c:n},\mathcal{V}_c)}}{\sum_{c=1}^C e^{cos(f^i_{c:n},\mathcal{V}_c)}},\label{eq:1}\vspace{-12pt}
\end{aligned}
\end{equation}}where $cos(\cdot)$ indicates the cosine similarity, and $C$ is the number of segmentation category. With the ground truth $b^i_c$, $\theta$ is trained with the binary cross entropy loss of 
{\begin{equation}\vspace{-10pt}
\begin{aligned}
   \mathcal{L}=-\sum_{c=1}^C \{c^i_{\mathbbm{1}}\log \sum_n^N p(b^i_{c:n},\mathcal{V}_c) + (1-c^i_{\mathbbm{1}})\log(1-\sum_n^N p(b^i_{c:n},\mathcal{V}_c))\},\label{eq:2}\vspace{-12pt}
\end{aligned}
\end{equation}}where $c^i_{\mathbbm{1}}$ is the one-hot class label. Therefore, the prompt refiner is optimized to be semantic aware for the specific anatomical structure. The prototype in buffer is also updated with the new model $\theta$. With $p(b^i_{c:n},\mathcal{V}_c)$ for each $b^i_{c:n}$, we pick the highest probability one  in $\mathcal{B}^i_c$ as our optimal proposal $b^i_{c:*}$ in current stage, which is also our bounding box inference.

Then, in the PGSR module, the off-the-shelf MedSAM~\cite{ma2024segment} takes both $x^i$ and $b^i_{c:*}$ to predict the possible segmentation mask $\hat{s}^i_c$. Notably, no information about the interested class can be directly informed in MedSAM. Although the MedSAM has strong ability of zero-shot segmentation to delineate the relatively accurate mask within the current box prompt $b^i_{c:*}$, the alignment of $\hat{s}_c^i$ is highly dependent on the precise $b^i_{c:*}$. This motivated us to refine $b^i_{c:*}$ with a more suitable proposal seed $b_{c:0}^i$ in SBPG. 

In practice, we can simply apply the smallest bounding box $\hat{b}^i_{c:0}$ for $\hat{s}^i_{c:0}$, and use $\hat{b}^i_{c:0}$ as our box proposal seed in the next round of SBPG. With $T$ rounds of iteration, we expect that the box-prompt can be progressively refined to inform the MedSAM to predict an accurate segmentation mask. 

In particular, in the last round of testing, we do not need to generate $\hat{b}^i_{c:0}$. Instead, the $T$-{th} round $\hat{s}^i_{c:0}$ is used as our final prediction.

\begin{table}[t!]
\centering  \vspace{-8pt}
\caption{Numerical comparisons and ablation studies of the cross-modality brain tumor HSI segmentation task}
\resizebox{0.9\linewidth}{!}{
\begin{tabular}{l|l|c|c}
\hline
Method & Supervision & Dice score $\uparrow$ & Hausdorff distance $\downarrow$\\\hline
WISE-Net& point &32.48 & 67.25 {[}mm{]}\\ 
\hline
Ours(T=1)& point&52.15 & 43.88 {[}mm{]}\\
Ours(T=5)& point&65.17 & 21.36 {[}mm{]}\\
Ours(T=10)& point&65.29 & 21.12 {[}mm{]}\\
\hline
MedSAM& box &68.74 & 18.42 {[}mm{]}\\
\hline
\end{tabular}\label{tab1}}  
\end{table}

\vspace{-8pt}
\section{Experiments and Results}
We evaluated on BraTS dataset for whole tumor segmentation as~\cite{liu2023incremental}, and use T2-weighted slices as our input. We used the 80\%/20\% split for training or testing. We generate the smallest bounding box for the mask and use its center as $\rho^i_c$. Our prompt refiner adopted the encoder part of the segmentor in~\cite{liu2023incremental}, which is followed by two fully connected layers of 1024 and 256 dimensions, respectively. We empirically set $X\times Y=21\times21$ and search proper $T\in\{1,2,3,5,10\}$. We followed~\cite{guo2023p2p} to compare with the conventional point-supervised method of WISE-Net without the use of the vision fundation model. Also, we directly input the ground truth of the bounding box into MedSAM, which can be ``upper bound" of the performance of pointwise method. In Tab.\ref{tab1}, we can see that our model with $T\geq5$ outperforms WISE-Net by a large margin. The Dice score is close to the box-supervised MedSAM.   

\vspace{-2pt}
\section{Conclusion}
In this work, we proposed an efficient iterative framework to enable box-prompt MedSAM to point-supervised medical image segmentation. A lightweight prompt refiner with only encoder is specifically trained for the interested structural class. We show that the fixed off-the-shelf large model can flexibly target the specific downstream task without the need of large scale fine-tuning. The medical PSS performance can be largely improved by the advance of vision foundation model, e.g., MedSAM, which show great promise for facilitating the image-guided interventions.

\vspace{-4pt} 
\section*{Acknowledgment}
This work is supported by R21EB034911, R01CA165221, R01DC018511, P41EB022544, and NAIRR240016. 

\bibliographystyle{ieee}
\bibliography{ref}

\end{document}